\documentclass[letterpaper, 10 pt, journal, twoside]{IEEEtran}

\usepackage{graphicx}
\usepackage{cite}
\usepackage[dvipsnames]{xcolor}
\usepackage{amsmath,amssymb,bm,mathtools}
\usepackage{tikz}
\usetikzlibrary{arrows.meta}
\usetikzlibrary{fit}
\usetikzlibrary{calc}
\usetikzlibrary{fadings}
\usepackage[urlcolor=blue]{hyperref}
\usepackage{multirow}
\usepackage{booktabs}

\newcommand\given[1][]{\:#1\vert\:}

\newcommand{\brm}[1]{\bm{\mathrm{#1}}}

\definecolor{myblue}{RGB}{0, 71, 255}
\definecolor{myred}{RGB}{255, 43, 43}
\definecolor{myyellow}{RGB}{255, 255, 100} 
\definecolor{mygreen}{RGB}{115,208,115} 
\definecolor{mygray}{RGB}{200, 200, 200}
\definecolor{myorange}{RGB}{234,134,66}
\definecolor{mypurple}{RGB}{50, 150, 255}

\tikzset{
    state/.style={
        draw,
        circle,
        fill=myblue,
        minimum size=0.16cm,
        inner sep=0
    },
    cosserat_factor/.style={
        draw,
        fill=myred,
        minimum size=0.13cm,
        inner sep=0,
        rectangle
    },
    yellow_factor/.style={
        draw,
        fill=myyellow,
        minimum size=0.13cm,
        inner sep=0,
        rectangle
    },
    orange_factor/.style={
        draw,
        fill=myorange,
        minimum size=0.13cm,
        inner sep=0,
        rectangle
    },
    prior_factor/.style={
        draw,
        fill=mygreen,
        minimum size=0.13cm,
        inner sep=0,
        rectangle
    },
    callout/.style={
        align=center,
        draw,
        rounded corners=1pt,
        line width=0.25,
        inner sep=2pt,
    },
    callout arrow/.style={
        -{Stealth[length=1.5mm,width=1.5mm]},
        line width=0.7pt,
        draw=mypurple
    },
}

\begin{document}

\title{
Continuum Robot State Estimation with \\ Actuation Uncertainty
}

\author{
James~M.~Ferguson$^{1}$, Alan~Kuntz$^{1}$, and Tucker~Hermans$^{2}$%
\thanks{
Manuscript received: February 22, 2026; Revised May 7, 2026; Accepted May 28, 2026.
This paper was recommended for publication by Editor Javier Civera upon evaluation of the Associate Editor and Reviewers' comments.
}%
\thanks{
Research reported in this publication was supported by the Advanced Research Projects Agency for Health (ARPA-H) under the ALISS project, Award Number D24AC00415-00. 
The ARPA-H award of up to \$11,935,038 provided 100\% of the financial support for this work. 
The opinions and findings in this paper are solely the responsibility of the authors and do not necessarily represent the official views of ARPA-H.
}%
\thanks{
Source code, demo scripts, and numerical experiments are publicly available at
\href{https://github.com/Kuntz-Lab/bendier}{\tt \footnotesize github.com/Kuntz-Lab/bendier}.
}%
\thanks{
$^{1}$James Ferguson and Alan Kuntz are with the Department of Electrical and Computer Engineering and Department of Computer Science, Vanderbilt University, TN, USA.
{\tt\footnotesize james.m.ferguson@vanderbilt.edu}
}%
\thanks{
$^{2}$Tucker Hermans is with the Kahlert School of Computing and Robotics Center, University of Utah, USA and NVIDIA, Seattle, WA, USA.
}%
\thanks{Digital Object Identifier (DOI): see top of this page.}
}

\markboth{IEEE Robotics and Automation Letters. Preprint Version. Accepted May 2026.}
{Ferguson \MakeLowercase{\textit{et al.}}: Continuum Robot State Estimation with Actuation Uncertainty} 

\maketitle

\begin{abstract}

Continuum robots are flexible, slender manipulators well suited for confined surgical environments. 
In these settings, unknown interaction forces and model uncertainty significantly affect robot shape, motivating state estimation from external observations. 
Existing estimation methods either neglect actuation modeling or rely on simplified deterministic actuation models. 
In contrast, we jointly estimate robot shape, external loads, and actuation inputs using mechanically principled actuation priors. 
To achieve this, we present a discrete Cosserat rod formulation with piecewise-linear strain integration that provides high numerical accuracy while inducing a sparse factor graph structure for efficient nonlinear optimization. 
We extend the framework to tendon-driven and parallel robots in simulation and validate it experimentally on a surgical concentric tube robot. 
Overall, our approach enables principled real-time estimation across multiple robot architectures while providing direct access to manipulator Jacobians through the linearized factor graph.

\end{abstract}

\begin{IEEEkeywords}
Flexible Robotics, 
Probability and Statistical Methods, 
% SLAM,
Parallel Robots,
Optimization and Optimal Control
\end{IEEEkeywords}

\section{Introduction}
\label{sec:introduction}

\IEEEPARstart{C}{ontinuum} robots are a class of compliant manipulators whose backbones deform continuously in response to actuation and external loading. 
Their slender geometry enables operation in confined environments, making them ideal candidates for surgical applications \cite{burgner2015continuum}. 
The primary modeling objective is to predict backbone shape under actuation, which is classically formulated as a deterministic boundary value problem (BVP) in arclength, where the Cosserat rod equations impose hard mechanical constraints \cite{russo2023continuum}.

In practice, however, even high-fidelity models are affected by unmodeled effects, parameter uncertainty, and actuator variability.
Furthermore, continuum robots often exhibit ill-posed inverse problems \cite{rucker2011deflection, ferguson2024unified, aloi2022estimating}.
For example, in deflection-based force sensing, loads applied along stiff directions are only weakly observable, leading to ambiguous solutions \cite{rucker2011deflection}.
Such ambiguity is naturally represented through conditional distributions over robot state.
These challenges motivate probabilistic state-estimation approaches \cite{barfoot2024state, dellaert2021factor} that have been adapted from temporal dynamics to the spatial arclength domain \cite{anderson2017continuum, ferguson2024unified, lilge2022continuum, lilge2024state, lilge2025incorporating}.
Rather than enforcing mechanics as hard constraints, these methods estimate a conditional distribution over robot state variables given external observations, enabling uncertainty quantification in poorly constrained configurations.

However, existing approaches either do not model actuation or rely on deterministic actuation models.
To address this, we explicitly represent actuation inputs as random variables, enabling joint estimation of rod state and actuation while recovering posterior correlations useful for ill-posed configurations \cite{rucker2011deflection}.
This additionally provides a unified framework, rather than relying on a separate deterministic solution to construct an initial prior \cite{lilge2025incorporating}.
Finally, because actuation variables are jointly estimated with rod state, the resulting posterior correlations can also be exploited for simple manipulator Jacobian extraction (e.g., for trajectory tracking).

\tikzfading[name=fade to right,
  left color=transparent,
  right color=white]
\begin{figure}
    \centering
    \vspace{3pt}
    \input{figures/front_page.tikz}
    \vspace{-8pt}
    \caption{
    Experimental setup and example results with a surgical concentric tube robot.
    \textit{Top}: The method fuses uncertain actuation inputs (e.g., tube rotation/translation) and external observations (e.g., magnetic tracker) with a prior mechanics model.
    \textit{Bottom-left}: Given known external loads, the method estimates a posterior distribution over robot shape (red shaded regions).
    \textit{Bottom-right}: For unknown external loads, the method infers loading (violet: mean, gold: covariance, green: measured load) from backbone observations.
    }
    \label{fig:intro_figure}
    \vspace{-20pt}
\end{figure}

In this letter, we present a continuum robot estimation framework with probabilistic mechanical actuation models.
To simplify the coupling between robot shape and wrench variables, the continuous Cosserat rod model is discretized into a set of local constraints relating state variables.
A piecewise-linear strain model preserves numerical integration accuracy while maintaining a compact state representation.
We demonstrate accurate and efficient state estimation for tendon-driven and parallel continuum robots in simulation and additionally validate the approach experimentally on a surgical concentric tube robot.
Together, these results demonstrate a principled and efficient framework for continuum robot estimation across diverse actuation types.

Section \ref{sec:continuous_rod} reviews the Cosserat model, and Section \ref{sec:discrete_rod} presents the proposed discrete formulation. 
Sections \ref{sec:tendon_robot} and \ref{sec:parallel_robot} present tendon actuation and parallel robot extensions. 
Experimental validation is provided in Section \ref{sec:experiments}. 

\section{Related Work}

Many continuum robot estimation methods do not explicitly incorporate prior rod mechanics models, instead reconstructing backbone shape via geometric fitting from external sensing \cite{camarillo2008vision, hannan2005real, kim2014optimizing} or Fiber Bragg Grating (FBG) arrays \cite{modes2020shape, ryu2014fbg}.
Learning-based methods instead regress shape using neural networks \cite{zhao2021shape, kuntz2020learning}, achieving high accuracy within trained regimes.
While effective, these methods neglect mechanical priors that could improve accuracy.

To incorporate prior mechanics information, several works have fit rod models directly to data. 
Black et al. estimated tip forces in parallel robots \cite{black2017parallel}, and Aloi et al. inferred distributed loads in tendon-driven systems \cite{aloi2022estimating}.
However, these approaches lack principled uncertainty modeling.

Probabilistic methods (e.g., Kalman filtering \cite{brij2010ultrasound, borgstadt2015multi, qiao2021force}) directly address this limitation \cite{rucker2011deflection}. 
For instance, Anderson et al. modeled Cosserat rods as a stochastic process in arclength \cite{anderson2017continuum}, and Lilge et al. \cite{lilge2022continuum, lilge2024state} used Gaussian process regression with a Cosserat prior for state estimation. 
This was extended to incorporate actuation \cite{lilge2025incorporating} by modifying the prior mean with a deterministic point-moment tendon model. 
While this improved accuracy, the point-moment approximation neglects force–shape coupling and can be less accurate than fully coupled formulations \cite{rucker2011statics}. 
Ferguson et al. \cite{ferguson2024unified} proposed joint shape and load estimation, but still used a deterministic tendon model and was not real-time.

In contrast, we explicitly model force–shape coupling and represent actuation and wrench variables directly within the estimation graph. 
We demonstrate real-time estimation across tendon-driven, parallel, and concentric tube robots, whereas \cite{ferguson2024unified} considered simpler systems without real-time performance.

\begin{table}[t]
    \centering
    \caption{Estimated state variables in this work.}
    \label{tab:state_variables}
    \begin{tabular}{c c l l}

\toprule

\multirow{3}{*}{\shortstack{Base\\Cosserat\\states}}
& $\brm{T}_k$ & $\in \mathrm{SE}(3)$ & Material frame pose at node $k$ \\

& $\brm{\sigma}_k$ & $\in \mathbb{R}^6$ & Internal wrench/stress at node $k$ \\

& $\brm{f}_k$ & $\in \mathbb{R}^6$ & Total applied wrench at node $k$ \\

\midrule

\multirow{4}{*}{\shortstack{Extended\\continuum\\robot\\states}}
& $\brm{q}$ & $\in \mathbb{R}^N$ & Mechanical actuation inputs \\

& $\brm{f}^\mathrm{e}_k$ & $\in \mathbb{R}^6$ & External wrench at node $k$ \\

& $\brm{T}_p$ & $\in \mathrm{SE}(3)$ & Parallel robot platform pose \\

& $\brm{f}_p$ & $\in \mathbb{R}^6$ & External wrench acting on platform \\

\bottomrule
\end{tabular}

\vspace{2pt}
\begin{minipage}{0.95\linewidth}
\footnotesize
\textit{Note:} The total wrench $\brm{f}_k$ combines contributions from actuation $\brm{q}$ and external disturbances $\brm{f}^\mathrm{e}_k$. 
Without actuation, the model reduces to $\brm{f}_k = \brm{f}^\mathrm{e}_k$. 
While our formulation allows external disturbances at any $k$, our experiments consider only tip loading ($k=K$), leaving distributed external loading \cite{aloi2022estimating, ferguson2024unified} for future work.
\end{minipage}
    \vspace{-15pt}
\end{table}

\section{Continuous Cosserat Rod Model}
\label{sec:continuous_rod}

To model continuum robot backbones, we adopt the Cosserat rod formulation from \cite{lilge2022continuum}. 
We represent poses as $\brm{T} = (\brm{R}, \brm{t}) \in \mathrm{SE}(3)$, mapping body to spatial coordinates, and twists as $\brm{\varepsilon} = (\brm{\omega}, \brm{\nu}) \in \mathfrak{se}(3)$, ordered with rotational components first \cite{dellaert2012factor}.
We use the skew-symmetric cross-product operator $[\cdot]_\times$ defined such that $[\brm{w}]_\times \brm{v} = \brm{w} \times \brm{v}$. 
The corresponding Lie algebra ``hat'' operator is
$\brm{\varepsilon}^\wedge \in \mathfrak{se}(3)$.
The adjoint and Lie bracket operators are then expressed as
\begin{equation}
\mathrm{Ad}(\brm{T}) =
\begin{bmatrix}
\brm{R} & 0 \\
[\brm{t}]_\times \brm{R} & \brm{R}
\end{bmatrix},
\quad
\mathrm{ad}(\brm{\varepsilon}) =
\begin{bmatrix}
[\brm{\omega}]_\times & 0 \\
[\brm{\nu}]_\times & [\brm{\omega}]_\times
\end{bmatrix}.
\end{equation}

The Cosserat rod kinematics \cite{lilge2022continuum} prescribe that pose evolves spatially in arclength $s$ according to 
\begin{equation}
    \label{eq:cosserat_kinematics}
    \frac{d}{ds} \brm{T}(s) = \brm{T}(s)\brm{\varepsilon}(s)^\wedge,
    \quad
    \brm{\varepsilon}(s) =
    \begin{bmatrix}
        \brm{\omega}(s) \\
        \brm{\nu}(s)
    \end{bmatrix},
\end{equation}
where $\brm{\varepsilon}(s) \in \mathbb{R}^6$ is the body-frame generalized strain \cite{lilge2022continuum} with angular and linear components $\brm{\omega}(s)$ and $\brm{\nu}(s)$.
Under loading conditions, the strain  $\brm{\varepsilon}(s)$ deviates from its nominal, zero-load value $\bar{\brm{\varepsilon}}(s)$ according to the linear constitutive law
\begin{equation}
    \label{eq:linear_law}
    \brm{\sigma}(s) =
    \begin{bmatrix}
        \brm{m}(s) \\
        \brm{n}(s)
    \end{bmatrix}
    =
    \brm{\mathcal{K}}\big(\brm{\varepsilon}(s) - \bar{\brm{\varepsilon}}(s)\big),
\end{equation}
where the matrix $\brm{\mathcal{K}} \in \mathbb{R}^{6 \times 6}$ denotes the rod stiffness and $\brm{\sigma}(s) \in \mathbb{R}^6$ is the internal wrench (i.e., generalized stress \cite{lilge2022continuum}), composed of moment $\brm{m}$ and force $\brm{n}$. 

Lilge et al. \cite{lilge2022continuum} highlighted the similarity between Cosserat rod mechanics and classical rigid body dynamics. 
In our frame convention, this manifests as the internal wrench evolution 
\begin{equation}
    \label{eq:lilge_result}
    \frac{d}{ds} \brm{\sigma}(s) 
    = \mathrm{ad}(\brm{\varepsilon}(s))^T \brm{\sigma}(s) - \brm{f}(s)
    ,
\end{equation}
where $\brm{f}(s) \in \mathbb{R}^6$ indicates the distributed external wrench applied to the rod at $s$.
Direct stochastic integration is nontrivial \cite{lilge2022continuum} due to the nonlinear coupling between (\ref{eq:lilge_result}) and (\ref{eq:cosserat_kinematics}).

\section{Discrete Cosserat Rod Estimation}
\label{sec:discrete_rod}

To address the nonlinear coupling described above, we deviate from the continuous formulation \cite{lilge2022continuum, lilge2024state, ferguson2024unified, teetaert2025stochastic} by discretizing the arclength domain. 
While this sacrifices some analytical structure, it simplifies the inclusion of wrench and actuation variables and enables direct application of existing factor graph solvers \cite{dellaert2021factor}.

Thus, we represent the rod using discrete states at $K$ arclength nodes 
$\{s_k\}_{k=1}^K$. 
Each node contains the base variables listed in Table~\ref{tab:state_variables}.
These state variables are connected through residual factors that encode physical constraints and observations.
Together, the factors define a factor graph and corresponding joint likelihood over all states. 
Under Gaussian noise assumptions, this formulation yields a nonlinear least-squares objective.

\subsection{Kinematics Likelihood Factor}

A common approximation for the $\mathrm{SE}(3)$ kinematics (\ref{eq:cosserat_kinematics}) is \textit{piecewise-constant} strain along arclength \cite{barfoot2024state}. 
However, as shown in Section \ref{sec:map_estimation}, a \textit{piecewise-linear} approximation provides significantly higher numerical accuracy.
We therefore model strain as linearly varying over each interval $s \in [s_k, s_{k+1}]$, given endpoint strain values $\brm{\varepsilon}_k$ and $\brm{\varepsilon}_{k+1}$:
\begin{equation}
\brm{\varepsilon}(s) = \brm{\varepsilon}_k + 
(s - s_k) \Delta \brm{\varepsilon}_k / \Delta s, 
\quad
\Delta \brm{\varepsilon}_k = \brm{\varepsilon}_{k+1} - \brm{\varepsilon}_k,
\end{equation}
where $\brm{\varepsilon}_k = \brm{\mathcal{K}}^{-1}\brm{\sigma}_k + \bar{\brm{\varepsilon}}_k$ with $\brm{\sigma}_k = \brm{\sigma}(s_k)$ and $\bar{\brm{\varepsilon}}_k = \bar{\brm{\varepsilon}}(s_k)$, which follows from (\ref{eq:linear_law}).
We assume throughout that $\Delta s$ is uniform.
Under this interpolation, the kinematics admits a higher-order approximation via the Magnus expansion (see \cite{barfoot2024state}, Sec. 8.2), yielding
\begin{equation}
\label{eq:linear_strain}
\begin{gathered}
\brm{T}_{k+1} = \brm{T}_k \exp\!\left(\tilde{\brm{\varepsilon}}_k^{\wedge}\Delta s\right), 
\quad
\tilde{\brm{\varepsilon}}_k = \brm{\varepsilon}^\star_k + \brm{n}_{\varepsilon_k}, \\
\brm{\varepsilon}^\star_k =
\brm{\varepsilon}_k +
\frac{1}{2}\Delta \brm{\varepsilon}_k
- \frac{\Delta s}{12}\mathrm{ad}(\Delta \brm{\varepsilon}_k)\brm{\varepsilon}_k
+ \frac{\Delta s^2}{240}\mathrm{ad}(\Delta \brm{\varepsilon}_k)^2\brm{\varepsilon}_k.
\end{gathered}
\end{equation}
The first term recovers piecewise-constant strain, while the remaining terms correct for linear strain interpolation. 
The noise $\brm{n}_{\varepsilon_k} \sim \mathcal{N}(0,\brm{\Sigma}_\varepsilon)$ captures kinematic uncertainty such as deviations from the linear law (\ref{eq:linear_law}).
Note that all noise models in this work are assumed Gaussian, consistent with prior work; this may not capture more complex effects or outliers.

Rewriting (\ref{eq:linear_strain}) in residual form yields the kinematics likelihood factor
\begin{equation}
\label{eq:kinematics_factor}
J_{\varepsilon_k} =
\left\|\brm{e}_{\varepsilon_k}\right\|^2_{\brm{\Sigma}_\varepsilon^{-1}},
\quad
\brm{e}_{\varepsilon_k} =
\brm{\varepsilon}_k^\star -
\frac{1}{\Delta s}
\left(\ln(\brm{T}_k^{-1}\brm{T}_{k+1})\right)^\vee ,
\end{equation}
which penalizes local deviations from ideal elasticity theory.
Note that for nonlinear optimization, each Gaussian factor in this work requires both a residual (e.g., (\ref{eq:kinematics_factor})) and Jacobian definition.
See Appendix \ref{sec:linearization} for linearization details.

\subsection{Mechanics Likelihood Factor}

\begin{figure}
    \centering
    \begin{tikzpicture}[scale=1.0, every node/.style={font=\footnotesize}]
    \def\dy{0.9}
    \def\dx{0.8}

    \tikzset{
        T_label/.style={right, xshift=2pt, yshift=-2pt},
        S_label/.style={anchor=south, yshift=2pt},
        F_label/.style={right, xshift=2pt}
    }

    \node[state] (T1) at (0,0) {};
    \node[T_label] at (T1) {$\brm{T}_1$};
    
    \node[state] (S1) at (0,\dy) {};
    \node[S_label] at (S1) {$\brm{\sigma}_1$};
    
    \node[state] (F1) at (0,2*\dy) {};
    \node[F_label] at (F1) {$\brm{f}_1$};

    \node[cosserat_factor] (E1) at (\dx,\dy) {};

    \node[state] (T2) at (2*\dx,0) {};
    \node[T_label] at (T2) {$\brm{T}_2$};
    
    \node[state] (S2) at (2*\dx,\dy) {};
    \node[S_label] at (S2) {$\brm{\sigma}_2$};
    
    \node[state] (F2) at (2*\dx,2*\dy) {};
    \node[F_label] at (F2) {$\brm{f}_2$};
    
    \draw (T1) -- (E1);
    \draw (S1) -- (E1);
    \draw (T2) -- (E1);
    \draw (S2) -- (E1);
    \draw (F2) -- (E1);

    \node[yellow_factor] (EBase) at (-\dx,\dy) {};

    \draw (T1) -- (EBase);
    \draw (S1) -- (EBase);
    \draw (F1) -- (EBase);
    
    \node[state] (T3) at (4*\dx,0) {};
    \node[T_label] at (T3) {$\brm{T}_3$};
    
    \node[state] (S3) at (4*\dx,\dy) {};
    \node[S_label] at (S3) {$\brm{\sigma}_3$};
    
    \node[state] (F3) at (4*\dx,2*\dy) {};
    \node[F_label] at (F3) {$\brm{f}_3$};

    \node[cosserat_factor] (E2) at (3*\dx,\dy) {};

    \draw (T2) -- (E2);
    \draw (S2) -- (E2);
    \draw (T3) -- (E2);
    \draw (S3) -- (E2);
    \draw (F3) -- (E2);

    \node[cosserat_factor] (E3) at (5*\dx,\dy) {};

    \draw (T3) -- (E3);
    \draw (S3) -- (E3);
    
    \node[state] (TK) at (7*\dx,0) {};
    \node[T_label] at (TK) {$\brm{T}_K$};
    
    \node[state] (SK) at (7*\dx,\dy) {};
    \node[S_label] at (SK) {$\brm{\sigma}_K$};
    
    \node[state] (FK) at (7*\dx,2*\dy) {};
    \node[F_label] at (FK) {$\brm{f}_K$};

    \node[cosserat_factor] (EKm1) at (6*\dx,\dy) {};

    \draw (TK) -- (EKm1);
    \draw (SK) -- (EKm1);
    \draw (FK) -- (EKm1);

    \draw[dotted, line width=1.5] ($(E3)!0.3!(EKm1)$) -- ($(E3)!0.7!(EKm1)$);
    
    \node[yellow_factor] (ETip) at (8*\dx,\dy) {};
    
    \draw (TK) -- (ETip);
    \draw (SK) -- (ETip);
    \draw (FK) -- (ETip);

    \path (T1) ++(-\dx,0) coordinate (T1PriorCoord);
    \node[prior_factor] (T1Prior) at (T1PriorCoord) {};
    \draw (T1) -- (T1Prior);

    \path (F1) ++(-0.3,0.3) coordinate (F1PriorCoord);
    \node[prior_factor] (F1Prior) at (F1PriorCoord) {};
    \draw (F1) -- (F1Prior);

    \path (F2) ++(-0.3,0.3) coordinate (F2PriorCoord);
    \node[prior_factor] (F2Prior) at (F2PriorCoord) {};
    \draw (F2) -- (F2Prior);

    \path (F3) ++(-0.3,0.3) coordinate (F3PriorCoord);
    \node[prior_factor] (F3Prior) at (F3PriorCoord) {};
    \draw (F3) -- (F3Prior);

    \path (FK) ++(-0.3,0.3) coordinate (FKPriorCoord);
    \node[prior_factor] (FKPrior) at (FKPriorCoord) {};
    \draw (FK) -- (FKPrior);
\end{tikzpicture}
    \vspace{-5pt}
    \caption{Factor graph representation of the discrete Cosserat rod model. Cosserat factors (red) encode physics constraints (\ref{eq:kinematics_factor}), (\ref{eq:mechanics_factor}) between rod state variables (blue). When actuation information is not available, external wrench prior factors (\ref{eq:external_wrench_prior}) can be specified (green). Boundary factors (\ref{eq:boundary_factors}) enforce consistency between the tip loads and internal wrenches (yellow).}
    \label{fig:cosserat_graph}
    \vspace{-15pt}
\end{figure}

To discretize the mechanics (\ref{eq:lilge_result}), we approximate external loading as a set of discrete wrenches applied at the $K$ node locations. 
This representation is exact for inherently discrete loads (e.g., tendon routing discs) and can approximate distributed loads \cite{aloi2022estimating, ferguson2024unified} arbitrarily well with sufficiently fine discretization, which we defer to future work.

Instead of (\ref{eq:lilge_result}), we start with the spatial frame $(\cdot)^w$ balance (see \cite{lilge2022continuum}) and substitute the discrete external wrenches $\brm{f}^w_k$ for the distributed load $\brm{f}^w(s)$:
\begin{equation}
    \frac{d}{ds} \brm{\sigma}^w(s) = -\brm{f}^w(s) = - \sum_{k=1}^K \brm{f}^w_k \, \delta(s - s_k)
    .
\end{equation}
The Dirac deltas $\delta(\cdot)$ imply $\frac{d}{ds} \brm{\sigma}^w(s) = 0$ on each interval $[s_k, s_{k+1}]$. 
Integrating across each impulse yields
\begin{equation}
    \label{eq:internal_wrench_world}
    \brm{\sigma}^w_{k+1} = \brm{\sigma}^w_k - \brm{f}^w_{k+1}
    ,\quad 
    \brm{\sigma}^w_k = \brm{\sigma}^w(s_k)
    .
\end{equation}
We apply the adjoint wrench transport $\brm{\sigma}^w_k = \mathrm{Ad}^{-T}(\brm{T}_k)\brm{\sigma}_k$ to express all quantities in the body frame \cite{lilge2022continuum}. 
After simplification using adjoint identities and introducing Gaussian noise $\brm{n}_{\sigma_k} \sim \mathcal{N}(0, \brm{\Sigma}_\sigma)$, we obtain
\begin{equation}
    \label{eq:internal_wrench_body}
    \brm{\sigma}_{k + 1} =
    \mathrm{Ad}^T (\brm{T}_k^{-1} \brm{T}_{k+1}) \brm{\sigma}_k
    - \brm{\mathcal{R}}^T(\brm{T}_{k+1}) \brm{f}_{k+1}
    + \brm{n}_{\sigma_k},
\end{equation}
where $\brm{\mathcal{R}}(\brm{T}_k) =
\begin{bmatrix}
\brm{R}_k & 0 \\
0 & \brm{R}_k
\end{bmatrix}$.
We define $\brm{f}_k$ as the body-frame external wrench expressed in spatial coordinates such that
$\brm{\mathcal{R}}^T(\brm{T}_k)\brm{f}_k = \mathrm{Ad}^{T}(\brm{T}_k)\brm{f}^w_k$, as is common \cite{rucker2011statics}.

Rearranging~\eqref{eq:internal_wrench_body} yields the mechanics residual
\begin{equation}
    \label{eq:mechanics_factor}
    \brm{e}_{\sigma_k} =
    \mathrm{Ad}^T (\brm{T}_k^{-1} \brm{T}_{k+1}) \brm{\sigma}_k
    - \brm{\mathcal{R}}^T(\brm{T}_{k+1}) \brm{f}_{k+1}
    - \brm{\sigma}_{k + 1},
\end{equation}
which is assumed Gaussian and therefore defines the likelihood factor given by $J_{\sigma_k} = \| \brm{e}_{\sigma_k} \|^2_{\brm{\Sigma}_{\sigma}^{-1}}$, penalizing deviations from ideal wrench propagation.
In practice, the noise $\brm{n}_{\sigma_k}$ is set small (e.g., $\brm{\Sigma}_\sigma = 10^{-6} \brm{I}_{6 \times 6}$) to softly enforce the wrench balance without imposing hard constraints.

\subsection{Boundary Values and Prior Factors}

\begin{figure}
    \vspace{-10pt}
    \centering
    \includegraphics[width=0.5\linewidth, trim=0.3cm 0.1cm 0.2cm 0.3cm, clip]{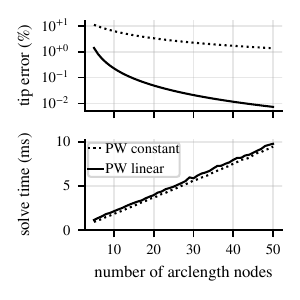}
    \hspace{40pt}
    \includegraphics[width=0.18\linewidth, trim=58cm 20cm 53cm 30cm, clip]{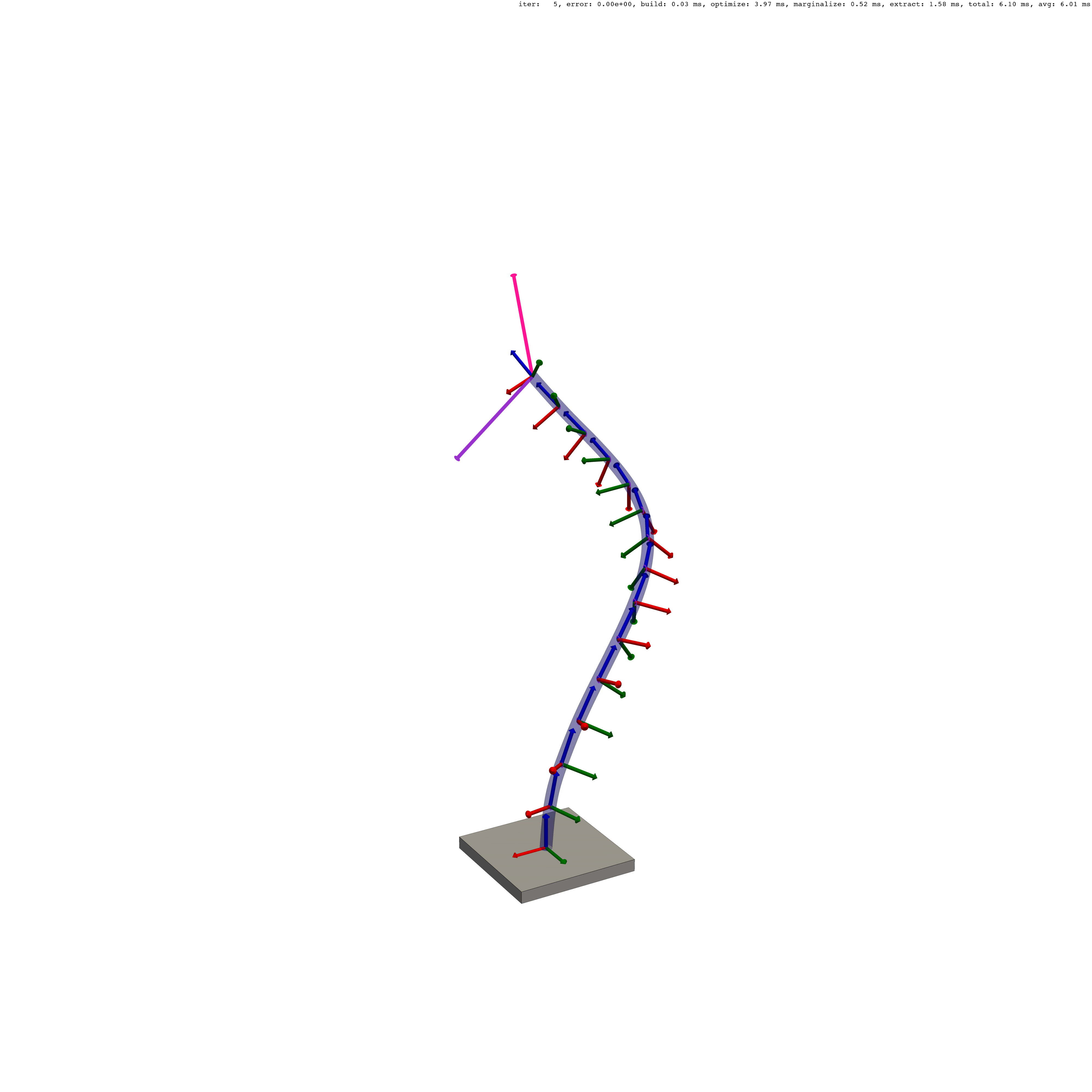}
    \vspace{-10pt}
    \caption{
    \textit{Top-left}: Average tip position error of our discrete Cosserat rod MAP solution relative to a BVP benchmark. Piecewise-linear and piecewise-constant strain integration rules are evaluated independently against BVP.
    \textit{Bottom-left}: Average solve time.
    \textit{Right}: Example configuration out of 540 distinct tip wrench conditions comprising force (violet) and moment (magenta) vectors.
    }
    \label{fig:cosserat_sim}
    \vspace{-15pt}
\end{figure}

At the base and tip of the rod, the internal wrenches $\brm{\sigma}_1, \brm{\sigma}_K$ must be caused solely by the respective external wrenches $\brm{f}_1, \brm{f}_K$.
This is enforced by the likelihood factors
\begin{equation}
\label{eq:boundary_factors}
\begin{aligned}
    J_{B_1} &= \left\| \brm{e}_{B_1} \right\|^2_{\brm{\Sigma}^{-1}_B}
    ,&
    \brm{e}_{B_1} &= \brm{\sigma}_1 + \brm{\mathcal{R}}^T(\brm{T}_1) \, \brm{f}_1
    ,\\
    J_{B_K} &= \left\| \brm{e}_{B_K} \right\|^2_{\brm{\Sigma}^{-1}_B}
    ,&
    \brm{e}_{B_K} &= \brm{\sigma}_K - \brm{\mathcal{R}}^T(\brm{T}_K) \, \brm{f}_K.
\end{aligned}
\end{equation}
We additionally anchor the base pose using the prior factor
\begin{equation}
    \label{eq:pose_prior}
    J_{T_1} = \left \| \brm{e}_{T_1} \right \|^2_{\brm{\Sigma}^{-1}_{T_1}}
    ,\quad 
    \brm{e}_{T_1} = 
    \ln
    \left(
        \bar{\brm{T}}_1^{-1} \brm{T}_1
    \right)^\vee
    ,
\end{equation}
which penalizes deviations of $\brm{T}_1$ from its prior mean $\bar{\brm{T}}_1$.

To solve the system, additional factors need to be placed on the wrench variables $\brm{f}_k$. 
Under known loading conditions, this can be done by attaching prior factors to the wrench variables:
\begin{equation}
    \label{eq:external_wrench_prior}
    J_{f_k} = \left \| \brm{e}_{f_k} \right \|^2_{\brm{\Sigma}^{-1}_{f_k}}
    ,\quad 
    \brm{e}_{f_k} = \brm{f}_k - \bar{\brm{f}}_k 
    .
\end{equation}
In Section \ref{sec:tendon_robot}, we instead attach actuation factors for more realistic scenarios when $\brm{f}_k$ are not directly known.

\subsection{Sparse Nonlinear Optimization}
\label{sec:map_estimation}

Given all of the likelihood factors, the negative log probability is proportional to a sum of weighted squared residuals:
\begin{equation}
\label{eq:negative_log_likelihood}
\begin{gathered}
    J(\brm{T}_{1:K}, \brm{\sigma}_{1:K}, \brm{f}_{1:K}) 
    = J_{B_1} + J_{B_K} + J_{T_1}
    \\
    + \sum_{k = 1}^{K - 1} (J_{\varepsilon_k} + J_{\sigma_k})
    + \sum_{k = 1}^K J_{f_k}
    .
\end{gathered}
\end{equation}
Fig.~\ref{fig:cosserat_graph} gives the corresponding factor graph representation.
We reiterate that (\ref{eq:negative_log_likelihood}) describes a simple Cosserat rod under known loading and boundary conditions; later sections extend this formulation with additional actuation and measurement factors.
For brevity, we omit the full likelihood expressions in later sections in favor of factor graph representations.

We implement the graph in GTSAM \cite{dellaert2012factor} using the linearizations in Appendix \ref{sec:linearization}, optimizing with a dogleg optimizer and warm starting from the previous solution when possible.
We take the Laplace approximation at the MAP solution as the posterior.
GTSAM enables efficient extraction of marginals over arbitrary subsets of variables \cite{dellaert2021factor}. 
For poses, the marginal is represented by a mean $\bar{\brm{T}}_k \in \mathrm{SE}(3)$ with local perturbation $\brm{\xi}_k \sim \mathcal{N}(0,\brm{\Sigma}_{T_k T_k})$, such that $\brm{T}_k = \bar{\brm{T}}_k \exp(\brm{\xi}_k^\wedge)$.

To study discretization granularity, we performed a simulation study where we solved a Cosserat rod under many random tip loading conditions $\brm{f}_K$ using our method. 
We compared the MAP estimate to a baseline BVP solver to study numerical integration error, with results shown for different values of $K$ in Fig.~\ref{fig:cosserat_sim}. 
All experiments were run on a workstation with an AMD Ryzen Threadripper 7960X CPU and 128\,GB of RAM. 
Additional details are available in our open-source code \cite{bendier}.

\section{Tendon-Actuated Continuum Robots}
\label{sec:tendon_robot}

\begin{figure}
    \centering
    \begin{tikzpicture}[scale=1.0, every node/.style={font=\footnotesize}]
    
    \tikzset{
        disc_rectangle/.style={
            draw=none,
            fill=mygray,
            rectangle,
            inner xsep=0.07cm,
            inner ysep=0.4cm
        }
    }

    \def\tlist{0, 1/16, 2/16, 3/16, 4/16, 5/16, 6/16, 7/16, 8/16, 9/16, 10/16, 11/16, 12/16, 13/16, 14/16, 15/16, 1}
    \def\xscale{8.5}
    \def\yscale{0.55}
    
    \foreach \t [count=\i from 0] in \tlist {
        \pgfmathsetmacro{\ang}{180*\t} % degrees
        \pgfmathsetmacro{\x}{\t * \xscale}
        \pgfmathsetmacro{\y}{(1 - cos(\ang))*\yscale} % degrees
        \coordinate (X\i) at ($(\x,\y)$);
    }

    % Ground plane
    \node[draw=none, fill=mygray, rectangle, inner xsep=0.1cm, inner ysep=1.0cm] at (X0) {};

    % Discs
    \node[disc_rectangle] at (X4) {};
    \node[disc_rectangle] at (X8) {};
    \node[disc_rectangle] at (X12) {};
    \node[disc_rectangle] at (X16) {};
    
    \def\XCoords{}
    \foreach \i in {0,...,16}{
        \xdef\XCoords{\XCoords (X\i)}
    }

    \draw[smooth, mygray, line width=1mm] plot coordinates {\XCoords};
    \draw[smooth] plot coordinates {\XCoords};

    % Disc factors
    \node[yellow_factor] (D1) at ($(X4)+(0,0.9)$) {};
    \draw (X0) -- (D1);
    \draw (X4) -- (D1);
    \draw (X8) -- (D1);
    \node[yellow_factor] (D2) at ($(X8)+(0,0.9)$) {};
    \draw (X4) -- (D2);
    \draw (X8) -- (D2);
    \draw (X12) -- (D2);
    \node[yellow_factor] (D3) at ($(X12)+(0,0.9)$) {};
    \draw (X8) -- (D3);
    \draw (X12) -- (D3);
    \draw (X16) -- (D3);
    \node[yellow_factor] (D4) at ($(X16)+(0,0.9)$) {};
    \draw (X12) -- (D4);
    \draw (X16) -- (D4);

    % Actuation prior connections
    \foreach \src [count=\i] in {4,3,2,1} {
        \path (D\src) ++(0, 0.4) coordinate (C);
        \draw (D\src) -- (C);
        \ifnum\i>1
            \draw (C_old) -- (C);
        \fi
        \coordinate (C_old) at (C);
    }

    \node[state] (tensions) at (C) {};
    \node[yshift=6pt] at (tensions.north) {$\brm{q}$};
    \path (tensions) ++(-0.5,0.5) coordinate (tensions_prior);
    \node[prior_factor] (C) at (tensions_prior) {};
    \draw (tensions) -- (C);
    \node[anchor=south, align=center] at (C.north) {actuation prior};
    
    % Force prior connections
    \foreach \src [count=\i] in {2,4,6,8,10,12,14,16} {
        \path (X\src) ++(0,-0.8) coordinate (C);
        \draw (X\src) -- (C);
        \ifnum\i>1
            \draw (C_old) -- (C);
        \fi
        \coordinate (C_old) at (C);
    }

    \node[prior_factor,
            label={[align=center, yshift=-1pt]below:
                   external \\ load \\ prior}] at ($(X15)+(0,-0.8)$) {};

    \def\L{6mm}

    \node[prior_factor,
          label={[align=center]left: measurements}]
          (M) at ($(X8)+(2.1,-1.3)$) {};

    \foreach \ang in {130, 90, 50}{
        \coordinate (mid) at ($(M)+(\ang:0.5*\L)$);
        \coordinate (end) at ($(M)+(\ang:\L)$);
    
        \draw (M) -- (mid);                    % solid
        \draw[dotted, line width=0.3mm] (mid) -- (end);  % dotted tail
    }

    % Base pose prior
    \node[prior_factor] (C) at ($(X0)+(0.6,0.8)$) {};
    \node[anchor=south, align=center] at (C.north) {base \\ pose \\ prior};
    \draw (X0) -- (C);

    \node[state] at (X0) {};
    \node[cosserat_factor] at (X1) {};
    \node[state] at (X2) {};
    \node[cosserat_factor] at (X3) {};
    \node[state] at (X4) {};
    \node[cosserat_factor] at (X5) {};
    \node[state] at (X6) {};
    \node[cosserat_factor] at (X7) {};
    \node[state] at (X8) {};
    \node[cosserat_factor] at (X9) {};
    \node[state] at (X10) {};
    \node[cosserat_factor] at (X11) {};
    \node[state] at (X12) {};
    \node[cosserat_factor] at (X13) {};
    \node[state] at (X14) {};
    \node[cosserat_factor] at (X15) {};
    \node[state] at (X16) {};
\end{tikzpicture}
    \vspace{-17pt}
    \caption{Tendon robot factor graph showing only one node between each disc and only four discs for clarity. Nodes and factors along the backbone (red) are a condensed version of the Cosserat rod graph (Fig.~\ref{fig:cosserat_graph}). Actuation factors (yellow) model transmission of tendon tension into backbone wrenches via (\ref{eq:tendon_factor}). Given prior actuation and external loads, the graph performs forward kinematics with uncertainty; the inverse problem (external load estimation) is solved by attaching measurement factors to backbone states.}
    \label{fig:tendon_robot_graph}
    \vspace{-15pt}
\end{figure}

A priori knowledge of the backbone loads $\brm{f}_k$ is generally unavailable. 
Instead, actuator variables $\brm{q} \in \mathbb{R}^N$ (e.g., tendon tensions) must be mapped to $\brm{f}_k$ through an actuation model \cite{rucker2011statics}.
To account for actuator uncertainty, we model $\brm{q}$ as a random vector with a Gaussian prior factor
\begin{equation}
    \left\| \brm{q} - \bar{\brm{q}} \right\|^2_{\brm{\Sigma}^{-1}_{q}},
\end{equation}
where $\bar{\brm{q}}$ denotes the measured actuator values and $\brm{\Sigma}_{q}$ specifies the associated uncertainty.
We extend the base Cosserat graph (Fig.~\ref{fig:cosserat_graph}) by connecting $\brm{q}$ to the wrenches $\brm{f}_k$ through tendon actuation factors (Fig.~\ref{fig:tendon_robot_graph}).
The tendon-disc interaction physics is adapted from \cite{gao2017general} for state estimation.

We assume tendon routing discs are located at a subset of $D$ arclength nodes $\mathcal{D} = \{d_1, \ldots, d_{D}\} \subseteq \{1, \ldots, K\}$.
For non-disc nodes $k \notin \mathcal{D}$, the total wrench is $\brm{f}_k = \brm{f}^\mathrm{e}_k.$
For disc nodes $d \in \mathcal{D}$, we include tendon-disc interaction terms:
\begin{align}
\label{eq:disc_wrench_sum}
\brm{f}_d =
\sum_{i=1}^{N}
\left(\brm{f}^{-}_{d,i} + \brm{f}^{+}_{d,i}\right)
+ \brm{f}^\mathrm{e}_d + \brm{n}_{D_d}.
\end{align}
Here, $\brm{f}^{-}_{d,i}$ and $\brm{f}^{+}_{d,i}$ denote the wrenches induced by tendon $i$ on disc $d$ from the two adjacent discs (assumed zero at rod tips). 
The term $\brm{n}_{D_d} \sim \mathcal{N}(0,\brm{\Sigma}_D)$ captures small unmodeled effects such as tendon-disc friction and routing imperfections.

The wrenches $\brm{f}_{d,i}^j$, due to the previous or next disc $(j \in \{-, +\})$, are computed based on the $i^\text{th}$ tension/actuation variable $q_i$ and the relative routing disc hole locations:
\begin{equation}
\label{eq:disc_wrench_geometry}
\begin{gathered}
    \brm{f}_{d, i}^j = 
    \brm{\mathcal{R}}(\brm{T}_d)
    \begin{bmatrix} 
        \brm{h}_{d, i}^\wedge \\
        \brm{I}_{3\times3}
    \end{bmatrix}
    \tilde{\brm{f}}_{d, i}^j
    , \quad
    \tilde{\brm{f}}_{d, i}^j = q_i 
    \, 
    \frac{\brm{d}_{d, i}^j}{\| \brm{d}_{d, i}^j \|}
    ,\\
    \brm{d}_{d, i}^j = 
    \begin{bmatrix}
        \brm{I}_{3\times3} & 0    
    \end{bmatrix}
    \brm{T}_d^{-1} \brm{T}_{j} 
    \begin{bmatrix}
        \brm{h}_{j, i} \\
        1
    \end{bmatrix} 
    - \brm{h}_{d,i}
    .
\end{gathered}
\end{equation}
The normalized difference in (body frame) hole locations $\brm{h}_{d,i}$, $\brm{h}_{j,i}$ is taken as the direction of the tendon force, scaled by $q_i$.

Overall, we rearrange (\ref{eq:disc_wrench_sum}) to form another residual:
\begin{equation}
    \label{eq:tendon_factor}
    \brm{e}_{D_d} = 
    \sum_{i = 1}^{N}
    \left(
        \brm{f}_{d, i}^{-} + \brm{f}_{d, i}^{+}
    \right)
    + \brm{f}^\mathrm{e}_d 
    - \brm{f}_d 
    ,
\end{equation}
with associated likelihood factor $J_{D_d} =  \left \| \brm{e}_{D_d} \right \|^2_{\brm{\Sigma}^{-1}_{D}}$.
This corresponds to the yellow factors in Fig.~\ref{fig:tendon_robot_graph}, linking tension variables to backbone wrenches. 

\begin{table}[t]
    \centering
    \caption{Continuum robot simulation details.}
    \label{tab:simulation_results}
    \begin{tabular}{lcc}
\toprule
 & Tendon Robot & Parallel Robot \\
\midrule
Number of discrete nodes        & 33        & 15 per rod \\
Number of time steps            & 900       & 300 \\
Mean baseline tip error$^\dagger$ (\%)         & 0.89      & 0.00036 \\
Mean solve time (ms)            & 10.31      & 18.60 \\
\bottomrule
\end{tabular}

\vspace{0.25em}
\begin{minipage}{\linewidth}
\footnotesize
$^\dagger$ Mean tip position error between our method's MAP estimate and the BVP solver, normalized as a percentage of robot length. This reflects open-loop mechanics model integration error without tip measurements.
\end{minipage}
    \vspace{-15pt}
\end{table}
\begin{figure*}
    \centering
    \includegraphics[height=0.21\linewidth, angle=-2, trim=2cm 16cm 14cm 13cm, clip]{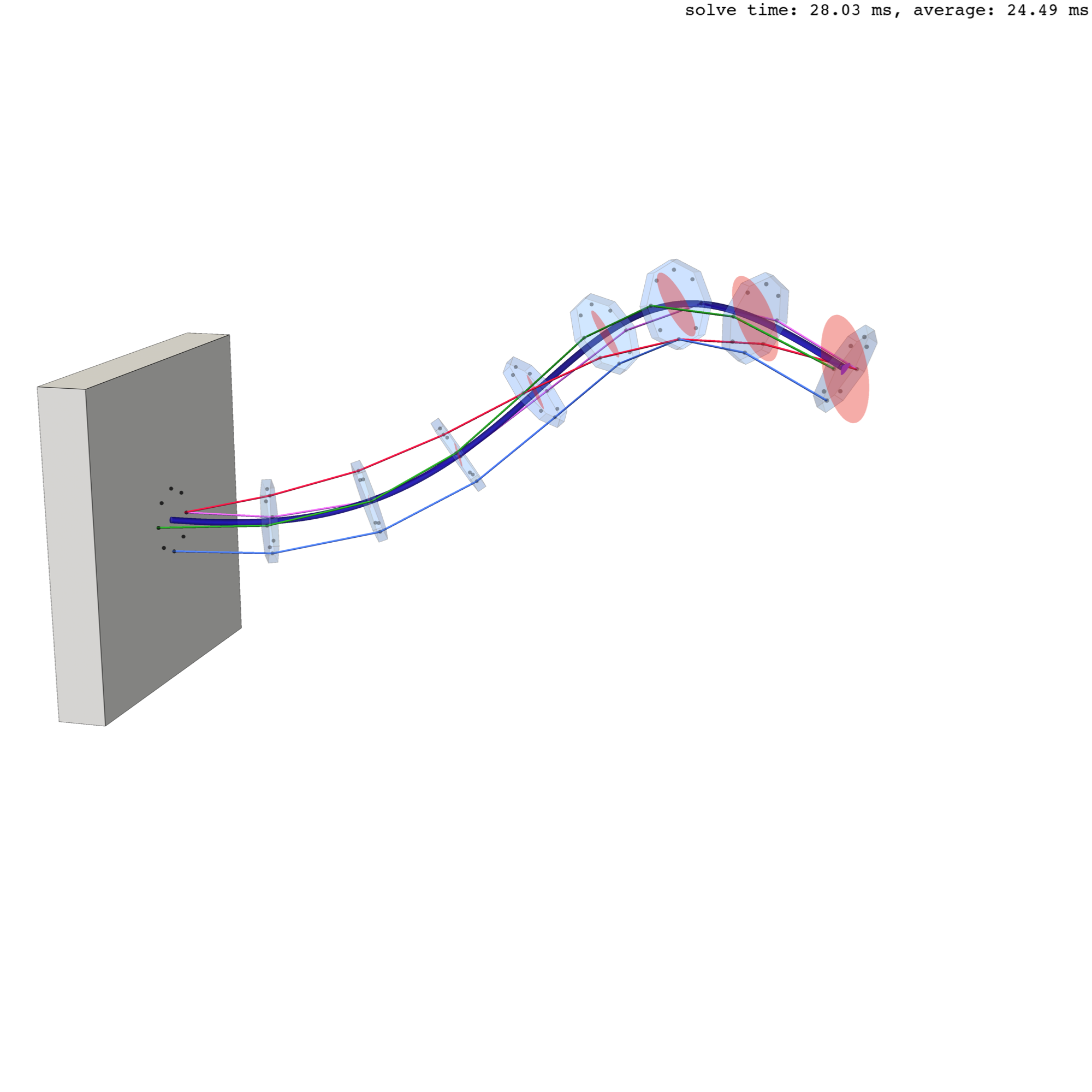}
    \hfill
    \includegraphics[height=0.21\linewidth, angle=-2, trim=2cm 16cm 14cm 13cm, clip]{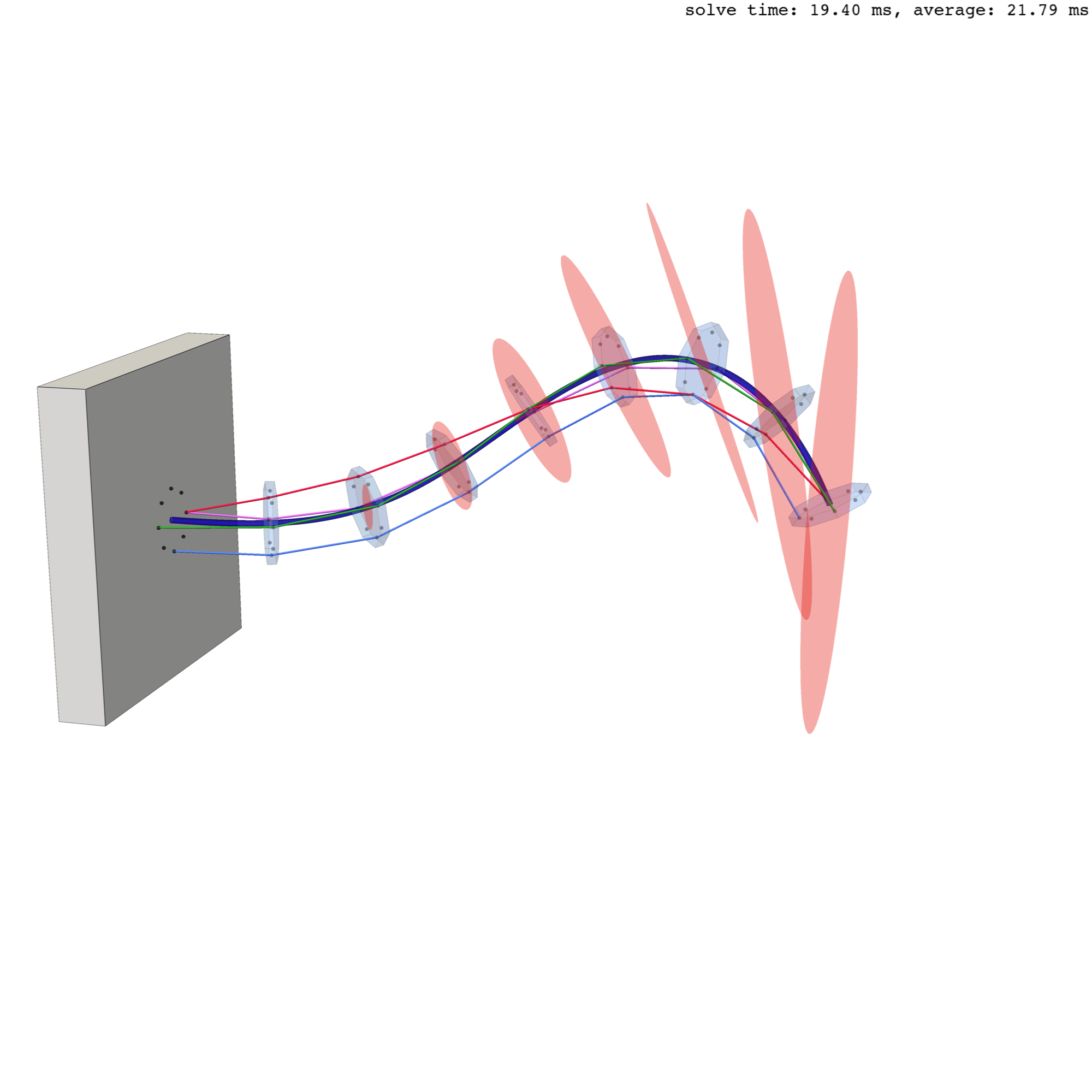}
    \hfill
    \includegraphics[height=0.21\linewidth, angle=-2, trim=2cm 16cm 20cm 13cm, clip]{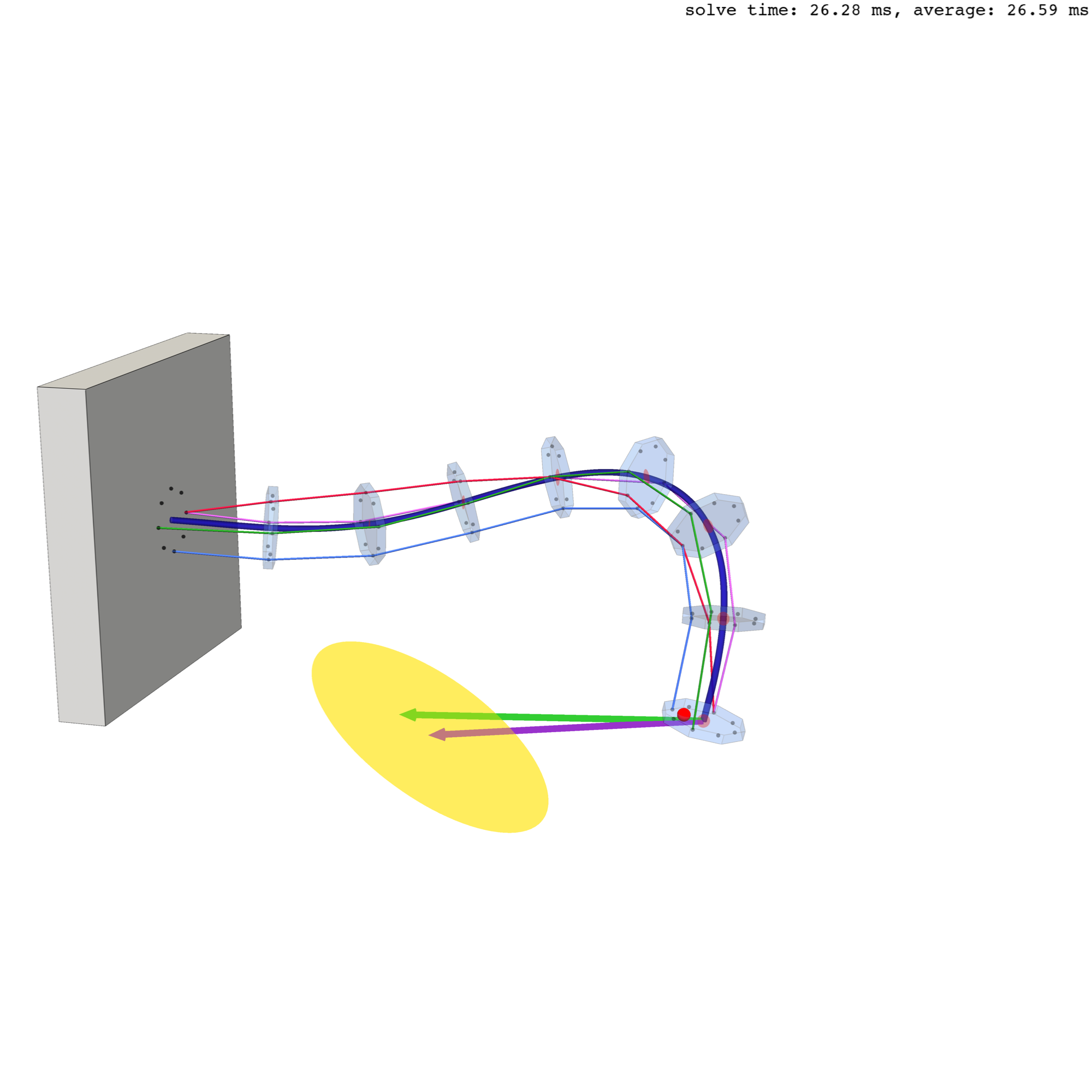}
    \vspace{-5pt}
    \caption{Example results from our tendon robot simulations illustrating different inference modes. \textit{Left}: Prediction with known (or zero) tip load $\brm{f}^\mathrm{e}_K$; red ellipsoids show shape uncertainty ($2\sigma$) along the backbone, subsampled at the disc locations. \textit{Middle}: Prior distribution under an unknown $\brm{f}^\mathrm{e}_K$ (zero mean, large prior covariance). \textit{Right}: Posterior distribution over $\brm{f}^\mathrm{e}_K$ given a noisy tip position observation (red point); violet and green arrows denote the MAP estimate and ground truth, with force uncertainty shown in gold.}
    \label{fig:tendon_robot_sim}
    \vspace{-20pt}
\end{figure*}

\begin{figure}
    \centering
    \includegraphics[width=\linewidth, trim=0.3cm 0cm 0cm 0cm, clip]{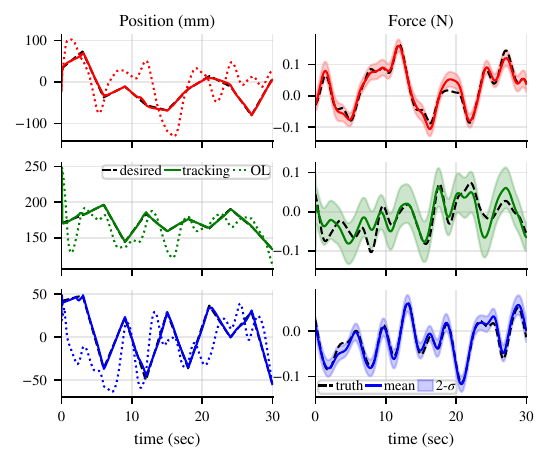}
    \vspace{-25pt}
    \caption{Tendon robot simulation results. 
    Red, green, and blue indicate $x, y,$ and $z$ components throughout.
    \textit{Left}: Tip position tracking in the presence of unknown loads. For comparison, we additionally show an open-loop (OL) trajectory without position feedback.
    \textit{Right}: Estimated tip forces while tracking, including $2\sigma$ uncertainty envelopes (shaded regions). The dotted lines indicate ground truth.}
    \label{fig:tendon_robot_results}
    \vspace{-15pt}
\end{figure}

The graph is optimized as in Section \ref{sec:map_estimation} and can be run in different modes depending on the chosen external wrench $\brm{f}^\mathrm{e}_k$ priors (see Fig.~\ref{fig:tendon_robot_sim}).
These priors are specified depending on the application and node location $k$. 
When $\brm{f}^\mathrm{e}_k$ is known (e.g., simulation ground-truth) or assumed zero (e.g., unloaded backbone regions), the prior mean is set accordingly with a small diagonal covariance (e.g., $10^{-6} \, \brm{I}_{6\times6}$), effectively constraining $\brm{f}^\mathrm{e}_k$. 
When $\brm{f}^\mathrm{e}_k$ is unknown, a zero-mean prior with large covariance is used, allowing inference from data. 
For tip force estimation (i.e., zero tip moment), the moment covariance entries are set small; e.g., {\tt blkdiag}$(10^{-6} \, \brm{I}_{3\times3}, 10^0 \, \brm{I}_{3\times3})$.

\subsection{Manipulator Jacobian Extraction}
\label{sec:jacobian_extraction}

Given the estimated state, we seek the manipulator Jacobian $\brm{J}_{T_K q}$ relating actuation to tip motions. 
Instead of finite differences \cite{black2017parallel} or sensitivity ODE integration \cite{rucker2011computing}, we extract $\brm{J}_{T_K q}$ directly from the linearized factor graph.

After optimization, the joint covariance $\brm{\Sigma}_{T_K q}$ (between tip pose $\brm{T}_K$ and actuation $\brm{q}$) and the marginal covariance $\brm{\Sigma}_{qq}$ immediately yield the best linear predictor of tip perturbation $\brm{\xi}_K$ given $\brm{q}$:
\begin{equation}
\label{eq:resolved_rates}
\begin{aligned}
    \mathbb{E}(\brm{\xi}_K \given \brm{q}) \approx
    \brm{\Sigma}_{T_K q} \, \brm{\Sigma}_{q q}^{-1} 
    \big( \brm{q} - \bar{\brm{q}} \big) =
    \brm{J}_{T_K q} \, \delta \brm{q}
    ,\\
    \implies
    \brm{J}_{T_K q} = \brm{\Sigma}_{T_K q} \, \brm{\Sigma}_{q q}^{-1}.
\end{aligned}
\end{equation}
This provides direct access to the manipulator Jacobian $\brm{J}_{T_K q}$ from the same linearization used for inference.
We demonstrate this approach in tip tracking simulations with both tendon-driven and parallel continuum robots.

\subsection{BVP Solver Benchmark Comparison}
\label{sec:tendon_benchmark}

With known tip wrench $\brm{f}^\mathrm{e}_K$, our approach should closely match standard BVP solutions.
To test this, we compare our MAP solution (e.g., Fig.~\ref{fig:tendon_robot_sim}, Left) with a solver based on \cite{gao2017general}. 
First, we use our Jacobian (\ref{eq:resolved_rates}) to generate a trajectory moving the tip position through a set of waypoints.
We separately sample a time-varying tip force $\brm{f}^\mathrm{e}_K$ from a 3D (zero moment) Gaussian process.
Given $\brm{q}$ and $\brm{f}^\mathrm{e}_K$, we solved for the MAP at each time step using both our approach and the BVP benchmark.
Full implementation details are publicly available at \cite{bendier}.
Results are shown in Table~\ref{tab:simulation_results}.

\subsection{Position Tracking with Tip Force Estimation}
\label{sec:tendon_sims}

In the previous section, unknown tip forces $\brm{f}^\mathrm{e}_K$ pushed the tip away from the desired trajectory. 
Here, we incorporate external measurements to estimate $\brm{f}^\mathrm{e}_K$ and enable disturbance rejection. 
When estimating loads, additional measurements are required to close the loop to ensure the problem is well-posed \cite{rucker2011deflection, ferguson2024unified}.
In particular, we use simple tip position measurements $\brm{z}_p$ for deflection-based force sensing \cite{rucker2011deflection}:
\begin{equation}
    \brm{z}_p = 
    \text{Pos} (\brm{T}_K)
    + \brm{n}_p
    , \quad 
    \brm{n}_p \sim \mathcal{N}(\mathbf{0}, \brm{\Sigma}_p)
    .
\end{equation}
The position $\brm{z}_p$ and actuation $\bar{\brm{q}}$ observations are gathered from a separate ground truth simulator with added Gaussian noise.
Given $\brm{z}_p$ and $\bar{\brm{q}}$, we estimate the state, including tip force $\brm{f}^\mathrm{e}_K$ (see Fig.~\ref{fig:tendon_robot_sim}, right).

\begin{figure}
    \centering
    \includegraphics[width=\linewidth, trim=0.3cm 0.1cm 0.5cm 0.3cm, clip]{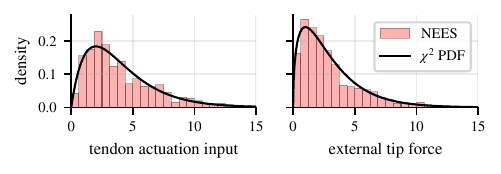}
    \vspace{-22pt}
    \caption{Normalized Estimation Error Squared (NEES) samples and theoretical distributions for tendon robot actuation (left) and external tip force estimation (right). Empirical 95\% confidence region coverage over 750 samples is 95.2\% and 94.7\%, respectively.}
    \label{fig:tendon_robot_nees}
    \vspace{-15pt}
\end{figure}

To track the desired trajectory, we use a damped velocity control law with the extracted Jacobian (\ref{eq:resolved_rates}) to update actuation commands. 
Tracking and estimation results are shown in Fig.~\ref{fig:tendon_robot_results}. 
We additionally ran a longer trajectory and recorded ground truth and estimated means and covariances for $\brm{q}$ and $\brm{f}^\mathrm{e}_K$. 
From these, we compute the Normalized Estimation Error Squared (NEES) and plot the samples as a histogram to assess estimator consistency in Fig.~\ref{fig:tendon_robot_nees}.
Full simulation details are available at \cite{bendier}.

\section{Parallel Continuum Robots}
\label{sec:parallel_robot}

Given the base Cosserat rod graph (Fig.~\ref{fig:cosserat_graph}), multiple rods can be coupled to form closed-loop topologies within our framework. 
A natural construction is to constrain the rod tips to a shared platform (Fig.~\ref{fig:parallel_robot_sim}). 
We present a factor graph formulation for the platform robot of \cite{black2017parallel}.
The approach is similar to \cite{lilge2024state}, but we also estimate wrenches.

Rather than enforcing platform constraints via an outer BVP solve \cite{black2017parallel}, we introduce constraint factors and optimize jointly. 
For rigid platform kinematics, we include factors relating each rod tip $\brm{T}_{K,i}$ to the platform pose $\brm{T}_p$:
\begin{equation}
\label{eq:parallel_relative_pose_factor}
\left\|
\ln\!\left(
\bar{\brm{T}}_{p,i}^{-1}\brm{T}_p^{-1}\brm{T}_{K,i}
\right)^\vee
\right\|^2_{\brm{\Sigma}^{-1}_{T_p}},
\end{equation}
where $\bar{\brm{T}}_{p,i}$ denotes the nominal tip pose offset of rod $i$.

To enforce the overall wrench balance between the rods and the platform, we again use the adjoint wrench transport on each of the six rod tips and sum to zero at $\brm{T}_p$:
\begin{equation}
    \label{eq:parallel_wrench_factor}
    \left \|
    \sum_{i = 1}^6 
    \mathrm{Ad}^T (\brm{T}_{K,i}^{-1} \brm{T}_p) \brm{\sigma}_{K,i}
    -    
    \brm{\mathcal{R}}^T (\brm{T}_p) \brm{f}_p
    \right \|^2_{\brm{\Sigma}^{-1}_{f_p}}
    .
\end{equation}
Finally, to incorporate linear actuation, we place vertical position displacement factors at the base of each rod:
\begin{equation}
    \Bigl(
    \begin{bmatrix}
        0 & 0 & 1
    \end{bmatrix}
    \mathrm{Pos}(\brm{T}_{1,i}) - \brm{q}_i
    \Bigr)^2 /  \sigma_{q_z}^2
    .
\end{equation}
We optimize the overall parallel robot factor graph in Fig.~\ref{fig:parallel_robot_sim}, in a similar way as before in Section \ref{sec:map_estimation}.

\subsection{Tip Force Sensing Simulations}

\begin{figure}
    \centering
    \begin{minipage}{0.44\linewidth}
        \begin{tikzpicture}[scale=1.0, every node/.style={font=\footnotesize}]
    % Define platform coords for rod termination and draw rectangle
    \coordinate (platform_center) at (0,0);
    \node[fill=mygray, draw=none, minimum width=2.3cm, minimum height=0.1cm, inner sep=0] at (platform_center) {};

    % End effector variables
    \path (platform_center) ++(0,1.5) coordinate (tip);
    \draw[mygray, line width=1mm] (platform_center) -- (tip);
    \node[state] (Ttip) at (tip) {};
    \node[xshift=-15pt] at (Ttip.west) {$\brm{T}_p, \brm{f}_p$};
    
    \path (platform_center) ++(-0.8,0) coordinate (platform_left);
    \path (platform_center) ++(0.8,0) coordinate (platform_right);

    % Define all coordinates along the rods
    \def\tlist{0, 0.33333, 0.66667, 1.0}

    \foreach \p/\xscale/\yscale/\name in {platform_left/-0.15/2.5/left, platform_center/-0.02/3.5/center, platform_right/0.15/3/right} {
        \foreach \t [count=\i from 0] in \tlist {
            \pgfmathsetmacro{\ang}{180*\t} % degrees
            \pgfmathsetmacro{\x}{(1 - cos(\ang))*\xscale} % degrees
            \pgfmathsetmacro{\y}{-\t*\yscale}
            \coordinate (\name\i) at ($(\p)+(\x,\y)$);
        }
    }

    % Draw rectangles at actuator basees
    \def\rectW{1cm}
    \def\rectH{0.1cm}
    \foreach \b in {left3,center3,right3}{
        \node[fill=mygray, draw=none, minimum width=\rectW, minimum height=\rectH, inner sep=0] at (\b) {};
    }
    
    % Draw the nodes/curves for the rods
    \foreach \name in {left,center,right}{
        \draw[smooth, mygray, line width=1.0mm] 
            plot coordinates {
                (\name0) (\name1) (\name2) (\name3)
            };
        \draw[smooth] 
            plot coordinates {
                (\name0) (\name1) (\name2) (\name3)
            };
        \node[state] (T\name0) at (\name0) {};
        \node[state] (T\name1) at (\name1) {};
        \node[state] (T\name2) at (\name2) {};
        \node[state] (T\name3) at (\name3) {};

        \node[cosserat_factor] at ($(T\name0)!.5!(T\name1)$) {};
        \node[cosserat_factor] at ($(T\name1)!.5!(T\name2)$) {};
        \node[cosserat_factor] at ($(T\name2)!.5!(T\name3)$) {};
    }

    \node[yellow_factor] (pose_constr_left) at ($(left0)!0.5!(tip)$) {};
    \node[anchor=east, align=center, xshift=-5pt] at (pose_constr_left) {platform \\ constraints};
    \draw (pose_constr_left) -- (Ttip);
    \draw (pose_constr_left) -- (Tleft0);
    
    \node[yellow_factor] (pose_constr_center) at ($(center0)!0.5!(tip)$) {};
    \draw (pose_constr_center) -- (Ttip);
    \draw (pose_constr_center) -- (Tcenter0);
    
    \node[yellow_factor] (pose_constr_right) at ($(right0)+(0,0.5*1.5)$) {};
    \draw (pose_constr_right) -- (Ttip);
    \draw (pose_constr_right) -- (Tright0);

    \node[orange_factor] (wrench_constr) at ($(right0)!0.4!(tip)$) {};
    \draw (wrench_constr) -- (Tleft0);
    \draw (wrench_constr) -- (Tcenter0);
    \draw (wrench_constr) -- (Tright0);
    \draw (wrench_constr) -- (Ttip);

    \node[prior_factor] (left_prior) at ($(left3)+(-0.4,0.4)$) {};
    \draw (Tleft3) -- (left_prior);
    
    \node[prior_factor] (center_prior) at ($(center3)+(-0.4,0.4)$) {};
    \draw (Tcenter3) -- (center_prior);
    \node[anchor=east, align=center, xshift=-3pt] at (center_prior) {actuator \\ values};
    \node[prior_factor] (right_prior) at ($(right3)+(0.4,0.4)$) {};
    \draw (Tright3) -- (right_prior);

    \node[prior_factor] (tip_prior) at ($(Ttip)+(0.5,0)$) {};
    \draw (Ttip) -- (tip_prior);
    \node[anchor=west, align=center, xshift=3pt] at (tip_prior) {wrench \\ prior};
\end{tikzpicture}
    \end{minipage}%
    \hspace{0.03\linewidth}%
    \begin{minipage}{0.5\linewidth}
        \centering
        \includegraphics[height=5cm, trim=44cm 15cm 52cm 65cm, clip]{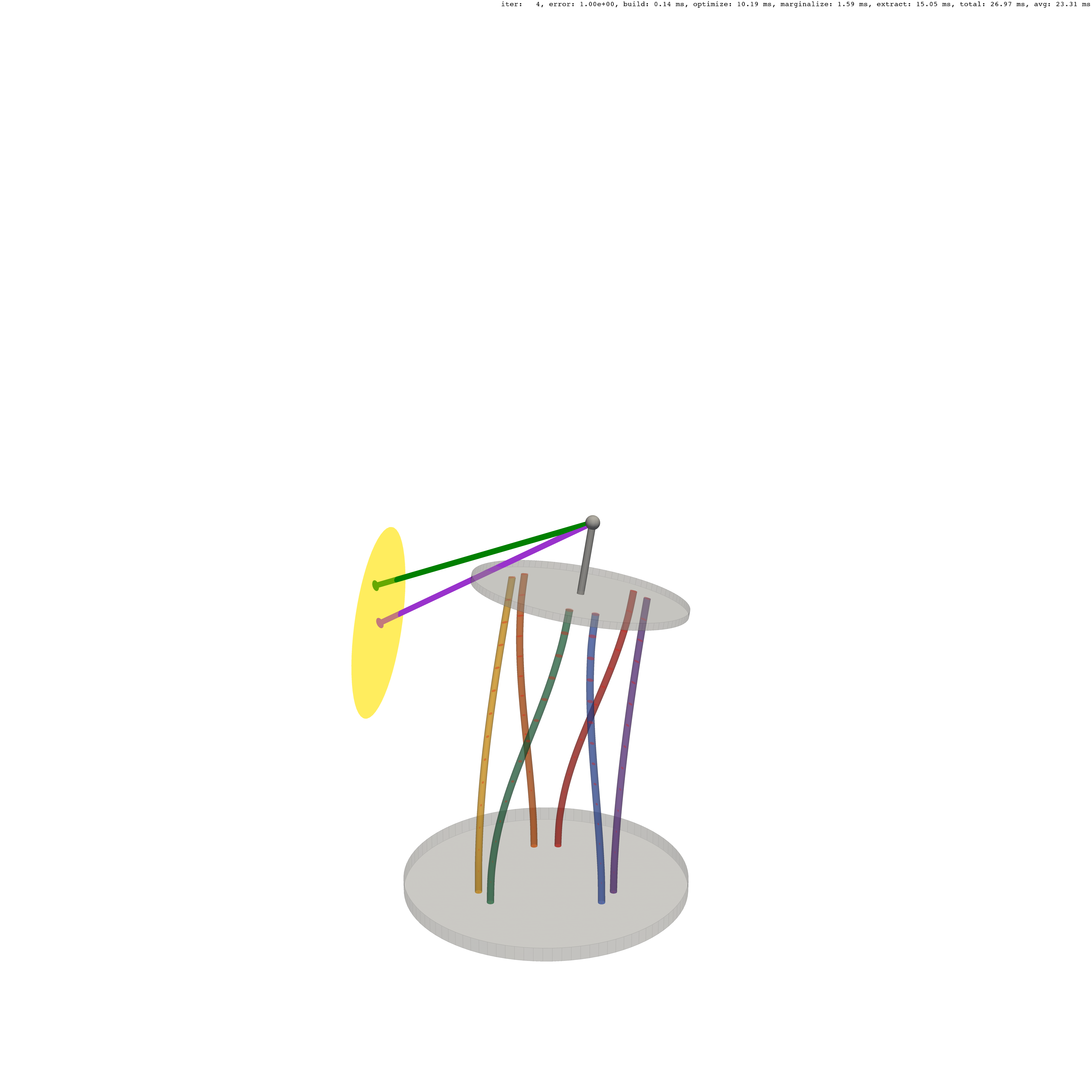}
    \end{minipage}
    \\
    \vspace{3pt}
    \includegraphics[width=\linewidth, trim=0.2cm 0cm 0.2cm 0cm, clip]{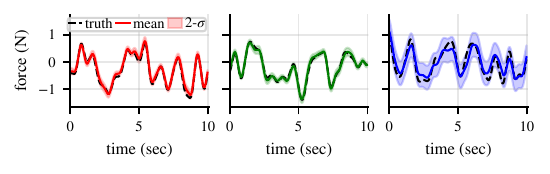}
    \vspace{-25pt}
    \caption{\textit{Left}: Parallel continuum robot factor graph with base linear actuation (see \cite{black2017parallel}).
    Multiple Cosserat graphs (see Fig.~\ref{fig:cosserat_graph}) are connected to a shared platform imposing relative pose factors (yellow) via (\ref{eq:parallel_relative_pose_factor}) and tip wrench factors (orange) via (\ref{eq:parallel_wrench_factor}).
    Priors from measured actuator positions/efforts constrain rod base states (green). 
    \textit{Right}: Example simulated estimation given actuator measurements.
    \textit{Bottom}: Estimated tip force over the trajectory.}
    \label{fig:parallel_robot_sim}
    \vspace{-15pt}
\end{figure}

Interaction forces on the platform can be inferred from linear actuator efforts at the rod bases \cite{black2017parallel}. 
In simulation, base vertical reaction forces are generated by a separate model with added noise. 
These measurements are incorporated via a simple prior wrench factor at each rod base:
\begin{equation}
\left(
\begin{bmatrix}
0 & 0 & 0 & 0 & 0 & 1
\end{bmatrix}
\brm{f}_{1,i} - \brm{z}_i
\right)^2 / \sigma_{f_z}^2 .
\end{equation}

We first evaluate numerical integration accuracy without measurements using a deterministic BVP baseline similar to \cite{black2017parallel} (Table~\ref{tab:simulation_results}).
For estimation with observations, we use the same simulation approach of Section \ref{sec:tendon_sims}, extracting the Jacobian via (\ref{eq:resolved_rates}) for tracking of a spiral-shaped reference trajectory. 
Full implementation details are publicly available at \cite{bendier}.
Example results are shown in Fig.~\ref{fig:parallel_robot_sim}.

\section{Concentric Tube Robot Experiments}
\label{sec:experiments}

Finally, we apply our framework to the Virtuoso endoscopic concentric tube robot (Fig.~\ref{fig:intro_figure}). 
Each arm consists of two telescoping Nitinol tubes ($\approx$1\,mm diameter, 40\,mm max length) actuated by independent base rotation and translation. 
The long endoscopic channel introduces torsion and backlash, leading to significant actuation uncertainty.

Since the inner tube is straight, we model the arm as two Cosserat rod graphs (see Fig.~\ref{fig:cosserat_graph}) in series.
The precurved outer tube is given a prior base pose (\ref{eq:pose_prior}) from its rotation angle; insertion actuation is enforced by an appropriate $\Delta s$.
The inner tube is set similarly by connecting its base pose to the outer tube tip pose.
Actuation uncertainty is incorporated by setting the covariance on these base poses.

For sensing, an NDI Aurora tracker coil is adhered at the tip, and interaction forces are measured using an ATI Nano17 force sensor via a silicone phantom (Fig.~\ref{fig:intro_figure}). 
All signals are recorded in ROS 2.

We first collect an initial dataset for calibration in which the robot is teleoperated throughout the workspace, palpating the phantom and exciting contact forces. 
A joint maximum-likelihood optimization is used to estimate force sensor alignment, tracker pose, tube stiffness, and noise parameters. 
A similar evaluation dataset is then used to test two operating modes. 
Representative results are shown in Fig.~\ref{fig:intro_figure}, with quantitative results provided in Fig.~\ref{fig:experiment_results}.

We additionally perform an ablation study comparing Mahalanobis error (i.e., $\sqrt{\mathrm{NEES}}$) with and without actuation noise included in the estimation model, in order to evaluate the impact of neglecting actuation uncertainty. 
Results are shown in Fig.~\ref{fig:experiment_input_noise}.

\section{Discussion}
\label{sec:discussion}

\subsection{Prediction Without Tip Observations}

Our approach produces MAP solutions closely matching BVP solvers, with tip position errors below $1\%$ of robot length in all cases (Table~\ref{tab:simulation_results}).
This accuracy is enabled by the piecewise-linear strain factor (\ref{eq:linear_strain}), which improves numerical performance over the typical piecewise-constant model, without a significant change in solve time (see Fig.~\ref{fig:cosserat_sim}).

We attribute remaining BVP errors to discretization and modeling differences (e.g., strain parameterization). 
This reflects a deliberate trade-off: unlike BVP solvers, our factor graph formulation prioritizes inference, modularity, and measurement integration. 
Sub-percent errors are likely negligible relative to modeling uncertainties (e.g., tendon friction).

\subsection{Posterior Given Tip Observations}

\begin{figure}
    \centering
    \includegraphics[width=\linewidth, trim=0.2cm 0cm 0.2cm 0cm, clip]{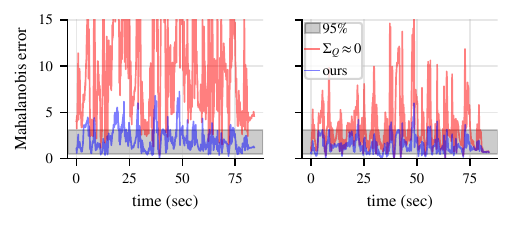}
    \vspace{-25pt}
    \caption{Experimental input noise analysis comparing the Mahalanobis distance between measured values and our estimate. We evaluate both with (blue) and without (red) actuation noise. \textit{Left}: Tip position given measured tip force. \textit{Right}: Tip force given measured tip position. Values exceeding the plot range are clipped.}
    \label{fig:experiment_input_noise}
    \vspace{-15pt}
\end{figure}

Across all experiments, estimation is consistent with ground truth (Figs.~\ref{fig:cosserat_sim}, \ref{fig:tendon_robot_results}, \ref{fig:parallel_robot_sim}, \ref{fig:experiment_results}), and NEES analysis confirms consistency with predicted uncertainty bounds (Fig.~\ref{fig:tendon_robot_nees}). 
Experimental results (Fig.~\ref{fig:experiment_results}) further validate the approach: given measured tip forces, the model accurately recovers robot shape, while the inverse force sensing problem also yields good results.
Fig.~\ref{fig:experiment_input_noise} further highlights the importance of modeling actuation uncertainty through reduced Mahalanobis error compared with the zero actuation noise case.

All solvers run at rates suitable for real-time quasi-static control (e.g., trajectory tracking), though no hard real-time guarantees are provided. 
In particular, the tendon robot solves in approximately 10\,ms versus approximately 4000\,ms in \cite{ferguson2024unified}, demonstrating the efficiency of the sparse optimization.

Jacobian extraction via (\ref{eq:resolved_rates}) is validated in tip tracking simulations: the estimated tip closely follows the reference trajectory even under unknown forces, whereas the open-loop model exhibits significant error (see Fig.~\ref{fig:tendon_robot_results}).
We note that errors in the estimated state can affect extracted Jacobians used for control (as is typical for manipulator Jacobians); however, this was not observed in our experiments.

Compared to continuous formulations \cite{ferguson2024unified, lilge2022continuum, lilge2024state, lilge2025incorporating}, our method does not use a Gaussian process prior and therefore lacks closed-form mean and covariance interpolation. 
However, sufficiently fine discretization is accurate in practice, and intermediate states could still be approximated via simple linear interpolation. 
Overall, our formulation can be viewed as an accurate discrete approximation, offering increased flexibility and modularity in the factor graph and simplifying incorporation of actuation and wrench variables for joint estimation.

\subsection{Ill-Posedness and Stability}

\begin{figure}
    \centering
    \includegraphics[width=\linewidth, trim=0.3cm 0cm 0cm 0cm, clip]{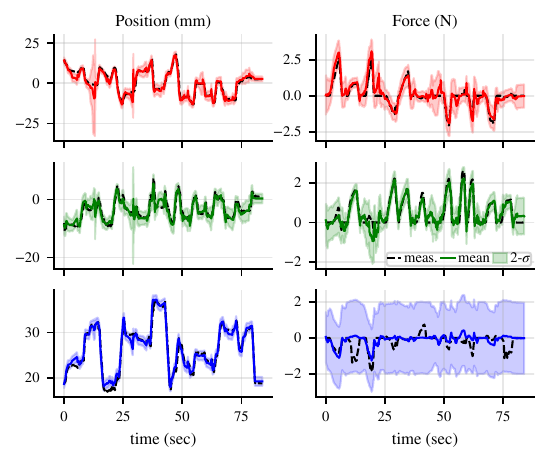}
    \vspace{-25pt}
    \caption{Experimental concentric tube robot estimation results.
    \textit{Left}: Tip position given measured tip forces (\textbf{2.23 ms} mean solve time over 900 time steps).
    Mean tip position accuracy relative to the tracker: \textbf{1.93 mm}. 
    \textit{Right}: Tip force given measured tip position (\textbf{2.33 ms} mean solve time).
    Mean tip force accuracy relative to the sensor: \textbf{0.44 N}.}
    \label{fig:experiment_results}
    \vspace{-15pt}
\end{figure}

Consistent with previous work \cite{rucker2011deflection}, in our experiments (Fig.~\ref{fig:experiment_results}), tip force sensing exhibits weak observability in the axial $z$ direction due to high stiffness relative to tracker noise \cite{rucker2011deflection}, resulting in estimates that remain close to the prior. 
Rather than becoming unstable, the estimator reflects this lack of information through posterior uncertainty, highlighting the need for additional sensing for full 3D force sensing.

More generally, this reflects inherent identifiability issues, where certain robot–sensing configurations yield ill-posed inverse problems \cite{rucker2011deflection, ferguson2024unified, aloi2022estimating}. 
For instance, in the tendon robot (Section \ref{sec:tendon_robot}), multiple combinations of actuation and external forces can produce indistinguishable tip observations. 
In our framework, these quantities are modeled as latent variables with priors, and when data is insufficient, ambiguity is represented in the posterior through strong correlations rather than a single arbitrary estimate.

Anecdotally, we further observed that some extreme configurations approach mechanically unstable regimes (e.g., buckling), consistent with prior work in continuum robot stability theory \cite{till2017elastic}.
Near unstable regimes, small parameter changes can produce large configuration changes, leading to snapping behavior.
Leveraging the graph structure to detect and avoid such configurations (e.g., through Hessian curvature analysis) is an important direction for future work.

\section{Conclusion}

We presented a framework for continuum robot state estimation with uncertain actuation models. 
Our discrete Cosserat rod formulation achieves high numerical integration accuracy and enables straightforward sparse optimization. 
We demonstrated the approach in simulation on tendon-driven and parallel continuum robots, validating against standard baselines, showing statistical consistency, and enabling direct extraction of manipulator Jacobians.
Experiments on a surgical concentric tube robot further validated estimation, with applications in tissue palpation. 
Overall, our results demonstrate a principled, real-time framework for continuum robot estimation across diverse actuation types.
An open-source implementation of the proposed framework is publicly available \cite{bendier}.

Future work will integrate additional sensing modalities (e.g., FBG arrays or endoscopic cameras) and extend the model to more complex systems (e.g., torsionally coupled concentric tubes). 
Given that our formulation supports external loads at arbitrary locations, a natural next step is estimating forces along the robot backbone \cite{aloi2022estimating, ferguson2024unified}. 
We further plan to extend the framework with space–time factors for temporal smoothing \cite{teetaert2025stochastic} or manipulator dynamics, and expect sparse solvers to achieve real-time smoothing in this setting.

\appendices

\section{Linearized Error Terms}
\label{sec:linearization}

Here we derive local linear approximations of our key residual models, which enable accurate and efficient factor graph optimization steps. 

Linearization of the first term $\brm{\varepsilon}_k^\star$ in (\ref{eq:kinematics_factor}) with respect to the strains $\brm{\varepsilon}_k, \brm{\varepsilon}_{k+1}$ is straightforward using adjoint properties, in particular, $\mathrm{ad}( \brm{x} ) \brm{y} = - \mathrm{ad}( \brm{y} ) \brm{x}$.
Full linearization with respect to the internal wrench variables $\brm{\sigma}_k, \brm{\sigma}_{k+1}$ comes from the linear law (\ref{eq:linear_law}).

Next we linearize the second term in (\ref{eq:kinematics_factor}) with respect to the pose variables, $\brm{T}_k = \bar{\brm{T}}_k \exp(\brm{\xi}_k^\wedge)$.
Defining $\bar{\brm{\Delta}} = \bar{\brm{T}}_k^{-1} \bar{\brm{T}}_{k+1}$, it is straightforward to show (Baker–Campbell–Hausdorff \cite{barfoot2024state}),
\begin{equation}
\label{eq:log_deriv}
\begin{gathered}
    \ln
    \left( \brm{T}^{-1}_k \brm{T}_{k+1} \right)^\vee
    =
    \ln
    \left( 
    \exp(-\brm{\xi}^\wedge_k)
    \bar{\brm{\Delta}}
    \exp(\brm{\xi}^\wedge_{k + 1})
    \right)^\vee
    \\
    \approx
    \ln
    \left( \bar{\brm{\Delta}} \right)^\vee
    -\brm{J}^{-1}_r (\bar{\brm{\Delta}}) \mathrm{Ad} (\bar{\brm{\Delta}}) \brm{\xi}_k + \brm{J}^{-1}_r (\bar{\brm{\Delta}}) \brm{\xi}_{k+1}
    ,
\end{gathered}
\end{equation}
where $\brm{J}_r$ is the right Jacobian of $\mathrm{SE}(3)$, which is available in closed form \cite{barfoot2024state}.

The mechanics error $\brm{e}_{\sigma_k}$ is only nonlinear in the pose variables.
For the first term in (\ref{eq:mechanics_factor}), we start by linearizing with respect to the intermediate variable $\brm{\Delta} = \brm{T}_k^{-1} \brm{T}_{k+1} = \bar{\brm{\Delta}} \exp(\brm{\delta}^\wedge)$:
\begin{equation}
\label{eq:adjoint_stuff}
\begin{gathered}
    \mathrm{Ad}^T 
    \left(
    \bar{\brm{\Delta}} 
    \exp \left(\brm{\delta}^\wedge \right) 
    \right)
    \brm{\sigma}_k
    \approx 
    \left(
    \brm{I}_{6\times6} + \mathrm{ad}^T (\brm{\delta})
    \right)
    \mathrm{Ad}^T \left(\bar{\brm{\Delta}} \right)
    \brm{\sigma}_k
    \\
    =
    \mathrm{Ad}^T \left( \bar{\brm{\Delta}} \right) \brm{\sigma}_k
    + \mathrm{ad}^T (\brm{\delta}) \mathrm{Ad}^T \left( \bar{\brm{\Delta}} \right) \brm{\sigma}_k
    .
\end{gathered}
\end{equation}
Letting $\mathrm{Ad}^T (\bar{\brm{\Delta}}) \brm{\sigma}_k = (\brm{\omega}, \brm{v})$, and $\brm{\delta} = (\brm{\delta}_\omega, \brm{\delta}_v)$, we rewrite the second term in (\ref{eq:adjoint_stuff}) as
\begin{equation}
    \begin{bmatrix}
    \brm{\delta}_\omega^\wedge & 0 \\
    \brm{\delta}_v^\wedge & \brm{\delta}_\omega^\wedge
    \end{bmatrix}^T
    \begin{bmatrix}
    \brm{\omega} \\
    \brm{v}
    \end{bmatrix}
    =
    \begin{bmatrix}
    \brm{\omega}^\wedge & \brm{v}^\wedge \\
    \brm{v}^\wedge & 0
    \end{bmatrix}
    \begin{bmatrix}
    \brm{\delta}_\omega \\
    \brm{\delta}_v
    \end{bmatrix}
    = \brm{\mathcal{A}} \, \brm{\delta}
    .
\end{equation}
Using this and the $\mathrm{SE}(3)$ perturbation linearization $\brm{\delta} = - \mathrm{Ad} (\bar{\brm{\Delta}}) \brm{\xi}_k + \brm{\xi}_{k+1}$, the overall linearization is 
\begin{equation}
\begin{gathered}
    \mathrm{Ad}^T 
    \left(
    \brm{T}_k^{-1} \brm{T}_{k+1}
    \right) 
    \brm{\sigma}_k
    \\
    \approx
    \mathrm{Ad}^T \left( \bar{\brm{\Delta}} \right) \brm{\sigma}_k 
    - \brm{\mathcal{A}} \, \mathrm{Ad} \left( \bar{\brm{\Delta}} \right) \brm{\xi}_k 
    + \brm{\mathcal{A}} \, \brm{\xi}_{k+1}
    .
\end{gathered}
\end{equation}
Linearizing the second term in (\ref{eq:mechanics_factor}) with respect to $\brm{T}_{k+1}$ follows from the $\mathrm{SO}(3)$ perturbation approximation
\begin{equation}
    \brm{R} \, \brm{v} = 
    \bar{\brm{R}} \exp( \delta \brm{r}^\wedge) \brm{v} 
    \approx 
    \bar{\brm{R}} \brm{v} - \bar{\brm{R}} \brm{v}^\wedge \delta \brm{r}
    .
\end{equation}

The derivatives of $\brm{e}_{D_d}$ in (\ref{eq:tendon_factor}) are obtained via repeated application of the sum, product, and chain rules, by linearizing (\ref{eq:disc_wrench_geometry}) with respect to the poses.
Other linearizations are either simple or straightforward extensions (e.g., (\ref{eq:parallel_wrench_factor}) is similar to (\ref{eq:mechanics_factor})). 
Implementation details for all of our factors, residuals, and derivatives are available at \cite{bendier}.

\bibliographystyle{IEEEtran}
\bibliography{library}

\end{document}